\documentclass[conference]{IEEEtran}
\IEEEoverridecommandlockouts
\usepackage{cite}
\usepackage{amsmath,amssymb,amsfonts}
\usepackage{algorithmic}
\usepackage{graphicx}
\usepackage{textcomp}
\usepackage{xcolor}
\usepackage{subfig}
\def\BibTeX{{\rm B\kern-.05em{\sc i\kern-.025em b}\kern-.08em
    T\kern-.1667em\lower.7ex\hbox{E}\kern-.125emX}}
    
\usepackage{numprint}

\def\RR{\mathbb{R}}

\begin{document}

\title{EWS-GCN: Edge Weight-Shared Graph Convolutional Network for Transactional Banking Data}

\author{\IEEEauthorblockN{Ivan Sukharev}
\IEEEauthorblockA{\textit{Risk Modeling \& Research, Sberbank} \\
Moscow, Russia \\
ivan@sukharev.me}
\and
\IEEEauthorblockN{Valentina Shumovskaia}
\IEEEauthorblockA{
\textit{Risk Modeling \& Research, Sberbank} \\
Moscow, Russia \\
valentinashumovskaya@gmail.com}
\and
\IEEEauthorblockN{Kirill Fedyanin}
\IEEEauthorblockA{\textit{Skolkovo Institute of Science and Technology}\\
Moscow, Russia \\
k.fedyanin@skoltech.ru}
\and
\IEEEauthorblockN{Maxim Panov}
\IEEEauthorblockA{\textit{Skolkovo Institute of Science and Technology (Skoltech)}\\
Moscow, Russia \\
m.panov@skoltech.ru}
\and
\IEEEauthorblockN{Dmitry Berestnev}
\IEEEauthorblockA{\textit{Risk Modeling \& Research, Sberbank} \\
Moscow, Russia \\
toberest@gmail.com}
}


\maketitle

\begin{abstract}
  In this paper, we discuss how modern deep learning approaches can be applied to the credit scoring of bank clients.  We show that information about connections between clients based on money transfers between them allows us to significantly improve the quality of credit scoring compared to the approaches using information about the target client solely. As a final solution, we develop a new graph neural network model EWS-GCN that combines ideas of graph convolutional and recurrent neural networks via attention mechanism. The resulting model allows for robust training and efficient processing of large-scale data. We also demonstrate that our model outperforms the state-of-the-art graph neural networks achieving excellent results.
\end{abstract}


\section{Introduction}
\label{sec:introduction}
  Risk management plays a core role in financial institutions, and banks actively invest in practices related to risk assessment. Currently, Machine Learning (ML) techniques became a standard instrument in banking for such tasks as client classification~\cite{Duan2009} and clustering~\cite{Zakrzewska2005}. Classical interpretable clients' assessment ML models usually require significant domain knowledge and laborious manual work to design features and the resulting model. Recently, financial institutions started to gradually switch to black-box models primarily based on Deep Learning (DL) approaches~\cite{Babaev2019,Leong2015,Zhu2018} which allow for efficient and automatic processing of banking data.

  A modern human, in particular a bank client, continually leaves traces in the digital world. For instance, the client may add information about transferring money to another person in a payment system. Therefore, every person obtains a large number of connections that can be represented as a directed graph. Such a graph gives an additional information for client's assessment. An efficient processing and usage of the rich heterogeneous information about the connections between clients is the main idea behind our study.

  In this work, we consider a graph, where nodes correspond to bank clients, while edges correspond to money transfers between them. As a target we consider the credit scoring problem for the bank clients, see more details in Section~\ref{sec:data}. In addition to transfers, for each client we use information about purchases, which was previously shown to be beneficial for classification of bank customers~\cite{Babaev2019}.
  We work under the assumption that the graph data can be turned into new valuable features for the model.
  However, in the considered case of transactional data the situation is very complex as we observe a multidimensional time series for each node and for each pair of nodes connected with an edge. The additional complication comes from the large size of the considered graph which has millions of nodes and billions of edges. 

  In this work, we consider models based only on the client purchases (for example, the Recurrent Neural Network (RNN)) as well as models explicitly employing the graph structure such as Graph Neural Network (GNN).

  The main contributions of the paper  are as follows.
  \begin{itemize}
    \item We show that an extensive usage of rich transactional data summarized as a graph with time series attributes allows to significantly improve the quality of credit scoring compared to the approaches based solely on the data for the target client.
    
    
    \item We propose a new GNN model \textit{EWS-GCN}, which uses both node and edge attributes employing special RNN as a feature encoder. The main peculiarity of the architecture is special attention mechanism combined with efficient weight-sharing scheme, which allows to significantly outperform state-of-the-art approaches in the experiments on the considered large-scale real-world banking dataset.
    
    \item We show how pretraining of submodels followed by an additional training of the whole pipeline allows to significantly improve the model quality compared to the training of the full model from random initialization.
  \end{itemize}


\section{Transactional Banking Data}
\label{sec:data}
  In this work, we consider the data provided by a large European bank. The data consists of clients' credit card operations, including both purchases and money transfers between clients. On top of that we are provided with clients' credit histories in this bank. The goal is to construct an efficient credit scoring model based on these data.

\subsection{Raw Data}
  Every transaction in the considered data is either an operation between individuals or an operation between an individual and a merchant. Each client is associated with a sequence of card operations, while a similar sequence of transfers corresponds to pairs of clients who have ever transferred money between them. 
  Descriptions of attributes used for purchases and transfers can be found in Table~\ref{tab:transactions}.

  \begin{table}[t]
  \centering
    \caption{The example of available information about purchases of a particular client and transfers between clients. For purchases additionally MCC codes are available.}
    \label{tab:transactions}
    \begin{tabular}{|l|ccc|}
      \hline
      Amount & 250 & 63 & 120 \\
      Currency & USD & EUR & EUR \\
      Date-time & 01/12/2019 22:31 & 02/12/2019 10:01 & 02/12/2019 13:47 \\
      \hline
    \end{tabular}
  \end{table}

  On top of the transactional data we are given the information about loan defaults. The loan default is an adverse event in which the client becomes unable to pay for a loan. We aim to predict the loan default defined as several consecutive non-payments within six months from the date of application, not paying attention to the further timeline.

\subsection{Graph Representation}
  Based on the raw data, we build a graph, which nodes correspond to bank clients, while edges correspond to interactions between them. We do not add specific nodes for merchants, since we aim to focus on the social connections. Instead, we add information about purchases to the nodes attributes as we think that purchases more characterize the node itself rather than its closest neighbors. Eventually, we build a large graph consisting of approximately a 100 million of nodes and about 10 billions of edges with rich transactional data available for both nodes and edges.

  To assess each particular client, we take only a subgraph around the target node of a depth two (that means that we take the client's neighbors and neighbors of neighbors). 
  It was motivated by two considerations:
  \begin{itemize}
    \item we aim to use the closest clients' connections, which can ``describe'' a person;
    
    \item the subgraph should fit into the memory, whereas in the considered graph the average degree of a node is close to a hundred and the subgraph size grows exponentially with the diameter of the neighborhood. 
  \end{itemize}
  Notice, that we do not consider the graph as a dynamic structure. Instead, we propose to build a graph based on the latest time segment (last year), since we aim to take into account exactly the latest clients activity in order to assess the risks related to the loan.

\section{Related Work}
\label{sec:related_work}
  The majority of the existing approaches for credit scoring task are based on aggregated transactional data, and classical machine learning models are applied~\cite{Chi2012, Khandani2010} as well as deep learning solutions~\cite{Bellotti2013}. In case of using of raw transactional data, in~\cite{Tobback2017} authors propose a model based on SVM classifier, while in~\cite{Babaev2019} RNN model is proposed. Recently, the idea to analyse bank clients as a part of a network was introduced in fraud detection problem~\cite{Tran2019} and in the task of embeddings construction~\cite{Bruss2019}, while in~\cite{Weber2018} authors solve a problem of anti-money laundering detection. 
  Our recent work~\cite{Shumovskaia2020} also considers bank clients as a network but focuses on a different task of finding stable connections between clients which is treated as a link prediction problem. 
  Talking about the latest DL developments in graph analysis, the standard GNN architectures frequently used in practice are GCN~\cite{Kipf2017}, GraphSAGE~\cite{Hamilton2017} and GAT~\cite{Velickovic2018}. 

  Note that in our case the additional information corresponding to edges of the graph is available, which is sequences of transfers with money amount and time stamps. To the best of our knowledge, edge features in GNNs are not well covered in the literature. Recently, EGNN~\cite{Gong2018} model is proposed which can handle edge features. However the detailed analysis reveals that the model has rather heavyweight architecture which needs a lot of data for training.

  Besides, we need to mention
  that we do not focus on the internal dynamics of a graph structure within the considered period period of time. Instead, we consider all transfers of the clients as edge attributes. Thus, we do not compare with neural network architectures for dynamic graphs like DCRNN~\cite{Li2018} or STGCN~\cite{Yu2018}.

\section{Edge Weight-Shared Graph Convolutional Network}
\label{sec:nn_main}
  In this section, we describe a new GNN architecture \textit{EWS-GCN}, which allows to aggregate information from both node and edge attributes. The distinctive properties of the model are special attention mechanism and the weight-sharing strategy for convolutional layers, which allow to build deeper architectures, outperforming existing models.

\subsection{Overview of Graph Convolutional Architectures}
\label{sec:standard_architectures}
  Let \(G = (X, E)\) be a subgraph extracted around a particular node, which we aim to classify. For the moment we will assume that we are given fixed-size representations for both nodes and edges. The generation of such embeddings from transactional data is discussed in Section~\ref{sec:rnn_encoder}. In what follows, we denote by 
  \begin{itemize}
    \item \(X\) -- a node feature matrix of size \(N \times d_x\), where \(N\) is the number of nodes, \(d_x\) is the dimension of the node feature vector;
    
    \item \(E\) -- a matrix of edge vector representations of size \(N \times N \times d_e\), where \(d_e\) is the dimension of edge feature vector.
  \end{itemize}

  The majority of existing GNN models perform a weighted aggregation of node features of the neighbors and differ in the algorithm for computing the weights. For example, GCN~\cite{Kipf2017} aggregates neighbors features normalizing them by the number of neighbors:
  \begin{equation*}
    \hat{x}_i = \sigma \biggl(\frac{1}{|\mathcal{N}_i|} \sum_{j \in \mathcal{N}_i} x_j W \biggr), ~ i = 1, \dots, N,
  \end{equation*}
  where \((x_1, \dots, x_N)\) are node embedding vectors before the convolution operation, \((\hat{x}_1, \dots, \hat{x}_N)\) are their counterparts after it, \(W\) are learnable weights, \(\mathcal{N}_i\) is a set of immediate neighbors of node \(i\) and, finally, \(\sigma\) is an activation function. The averaging operation implies that all the neighbors have an equal influence on the considered nodes which is apparently very unnatural in the majority of applications. 

  The latter issue was tackled in GAT~\cite{Velickovic2018} model, where neighbors' features are weighted according to the value of attention mechanism, computed based on node feature values:
  \begin{equation*}
    \hat{x}_i = \sigma \biggl(\mathop{\Vert}_{k = 1}^{K} \sum_{j \in \mathcal{N}_i} \alpha^k(x_i, x_j) \cdot \bigl(x_j W^k\bigr)\biggr), ~ i = 1, \dots, N,
  \end{equation*}
  where \(\Vert\) denotes concatenation, \(K\) is the number of channels, and there are separate learnable matrices \(W^k\) and functions \(\alpha^k(x_i, x_j)\) for each channel. The model becomes very flexible but with the price of sufficient increase in the number of parameters. Importantly, both GCN and GAT do not consider edge features in their approach.

  If some features \(e_{ij} \in \RR^{d_e}\) are available for edges they may be used to compute more accurate attention weights. The idea was implemented in quite complex EGNN~\cite{Gong2018} model:
  \begin{equation}
  \label{eq:egnn}
    \hat{x}_i = \sigma \biggl(\mathop{\Vert}_{k = 1}^{K} \sum_{j \in \mathcal{N}_i} \alpha^k(x_i, x_j, e_{ij}) \cdot \bigl(x_j W^k\bigr)\biggr), ~ i = 1, \dots, N.
  \end{equation}
  The authors of~\cite{Gong2018} limit the choice to the same matrix \(W^k \equiv W\) for all the channels and discuss different variants of constructing functions \(\alpha_k(x_i, x_j, e_{ij})\). 
  In this work, while staying almost in the generality of equation~\eqref{eq:egnn} we propose the particular instantiation well-adapted to the peculiarities of our data.

\subsection{Proposed Graph Convolutional Layer}
\label{sec:proposed_layer}
  \begin{figure*}[t!]
    \centering
    \includegraphics[width=420pt]{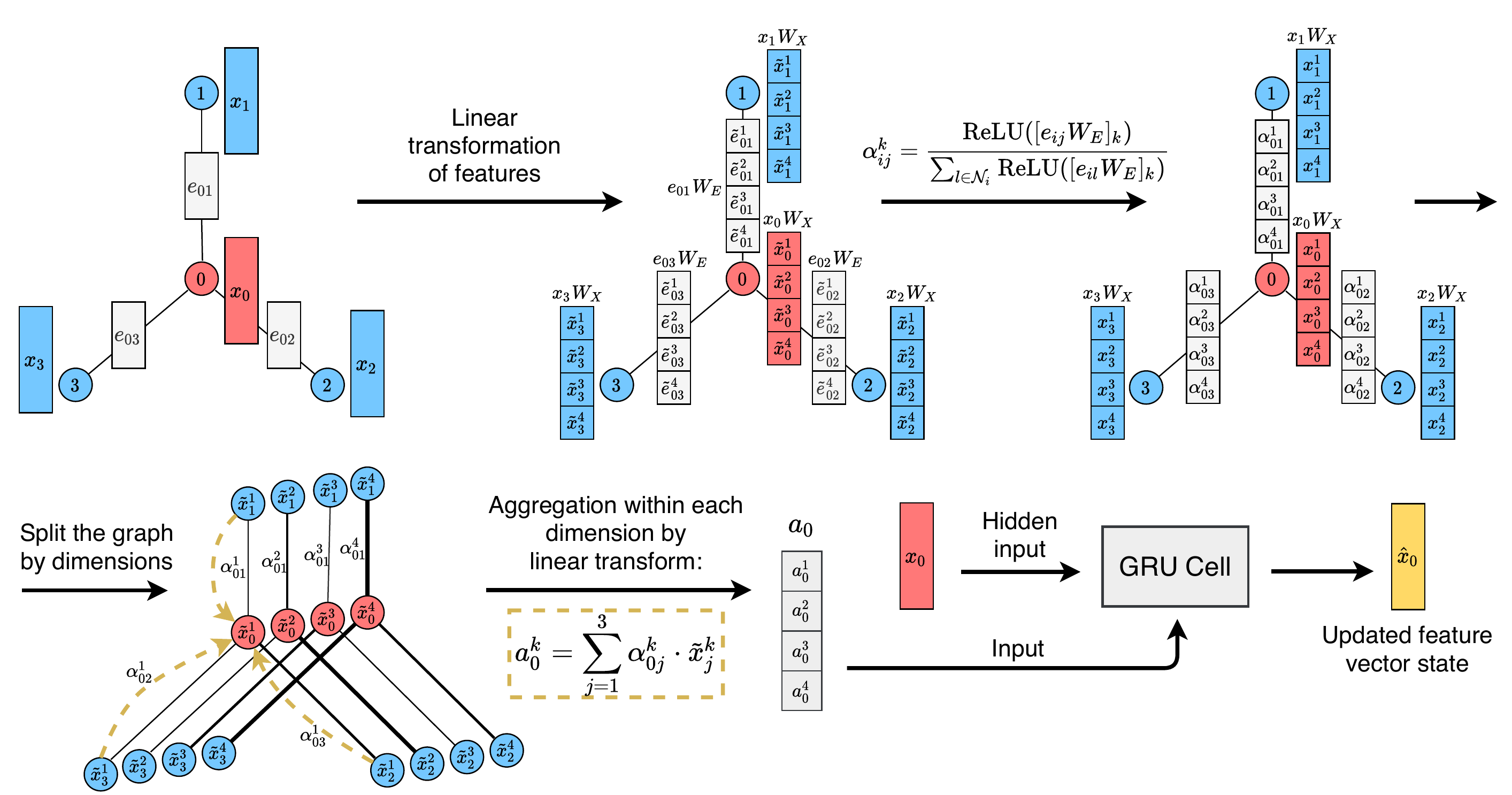}
    \caption{Convolutional layer scheme. The red node is the node we aim to classify, while blue ones are its neighbors. Initially, they have node features \(x_i\) and edge features \(e_{ij}\). Then, linear transformations are applied in order to get edge and node attributes of the same dimension. After that, we construct attention vectors \(\alpha_{ij}\) between the target node and its neighbors, based on edge features. The next step is ``splitting'' the graph by dimension (channels), and we aggregate the features within each dimension separately, obtaining final \(a_{0}\) vector. Finally, vector \(a_{0}\)  and the initial state \(x_0\) are passed to GRU cell. As a result, we get an updated feature vector.}
  \end{figure*}

  Our basic idea is to weight each node feature vector coordinate (channel) separately. We conduct a linear transformation of both node \(\tilde{x}_i = x_i W_X\) and edge \(\tilde{e}_{ij} = e_{ij} W_E\) features, where \(W_X \in \RR^{d_x \times K} \) and \(W_E \in \RR^{d_e \times K}\). Such transformation allows to achieve the same dimension \(K\) for both node and edge features.
  Then, we compute attention weights \(\alpha_{ij}^k\):
  \begin{equation*}
    \alpha_{ij}^k = \frac{\mathrm{ReLU}\bigl([e_{ij} W_E]_k\bigr)}{\sum_{l \in \mathcal{N}_i} \mathrm{ReLU}\bigl([e_{il} W_E]_k\bigr)}, ~ k = 1, \dots, K,
  \end{equation*}
  where \(\alpha\) is an attention tensor of size \(N \times N \times K\) and \([\cdot]_k\) denotes the \(k\)-th coordinate of the vector.
  There are 2 main motivations behind such a choice of attention function:
  \begin{enumerate}
    \item In our application, the main information about similarity of nodes is contained in edges (transfers between clients), while node features are less important. That is why our attention is based solely on edge features.
    
    \item \(\mathrm{ReLU}\) here produces sparsification by nullifying negative values, which allows to remove a part of neighborhood out of consideration based on the edge representation vector. This is an important step in speeding-up the computations and filtering out non-informative data.
  \end{enumerate}

  We use the obtained attention tensor \(\alpha\) to aggregate node features over the neighborhood:
  \begin{equation*}
    a_i = \sum_{j \in \mathcal{N}_i} \alpha_{ij} \odot (x_{j} W_X), ~ i = 1, \dots, N,
  \end{equation*}
  where \(\odot\) stands for element-wise multiplication and \(a_i \in \RR^{K}\) is an intermediate representation for node \(i\). 

  The final important step in every graph convolution is to update the node representation based on the current representation and linear combination of neighbours representations. The standard approach is just to apply elementwise nonlinearity, see Section~\ref{sec:standard_architectures}. However, it was observed that it is also beneficial to simply concatenate the new embedding and the one from the previous layer~\cite{Hamilton2017}. The main drawback of this approach is in growing dimension of the representation from layer to layer. In this work we suggest to consider non-linear learnable aggregation via GRU cell \(g\), similar to the GGNN model~\cite{Li2015}.  Thus, the final operation becomes:
  \begin{equation*}
    \hat{x}_i = g(a_i, x_i), ~ i = 1, \dots, N,
  \end{equation*}
  which persists the dimension of embedding: \(\hat{x}_i \in \RR^{d_x}\), while being flexible due to the learnability of the parameters of function \(g\). Combining all the steps together we obtain
  \begin{equation*}
    \hat{x}_i = g\biggl(\sum_{j \in \mathcal{N}_i} \alpha_{ij} \odot (x_{j} W_X), x_i\biggr), ~ i = 1, \dots, N.
  \end{equation*}
  %
  The proposed scheme is summarized on Figure~\ref{fig:convolution}. It is very flexible but has only few parameters and keeps the dimension of the node representation, thus allowing the recurrent usage. 

\subsection{Transactions and Transfers Encoder}
\label{sec:rnn_encoder}
  Graph convolutional models are limited to graphs with fixed-size feature vectors for both nodes and edges. In the considered case of transactional data, both types of attributes have much more complex structure of a time series with varying length. In order to fit the considered data into modern GNNs, we propose to process a series of transactions by RNN and use the embedding from one of its final layers as a feature vector. Importantly, we suggest to pretrain RNN on the same credit scoring problem, so that the obtained embeddings are meaningful from the prospective of the considered downstream task. 

  Let us define the time series of transactions for the client \(i\) as \(z_i\). Then the RNN is trained to solve credit scoring problem based on \(z_i\) for \(i = 1, \dots, N\), and corresponding target values \(y_i \in \{0, 1\}\) (not default -- default). As a result  for each node \(i\) we can compute an embedding vector \(x_i = f(z_i) \in \RR^{d_x}\), where function \(f(\cdot)\) outputs the last layer embedding of dimension \(d_x\) for trained RNN and given input.

  Importantly, in our problem transfers between clients have the same representation as transactions for clients which allows us to proceed them by the same pretrained RNN. Thus, for each pair of clients \(i\) and \(j\) with non-empty history of transactions \(z_{ij}\) we can compute \(e_{ij} = f(z_{ij})\). 

  The distinctive feature of our approach is that we can make an additional training of RNN models for nodes and edges together with the full GNN, which allows to further improve the quality of embeddings, see the implementation details in Section~\ref{sec:experiments}.

\subsection{EWS-GCN Model}
\begin{figure*}[t!]
    \centering
    \includegraphics[width=420pt]{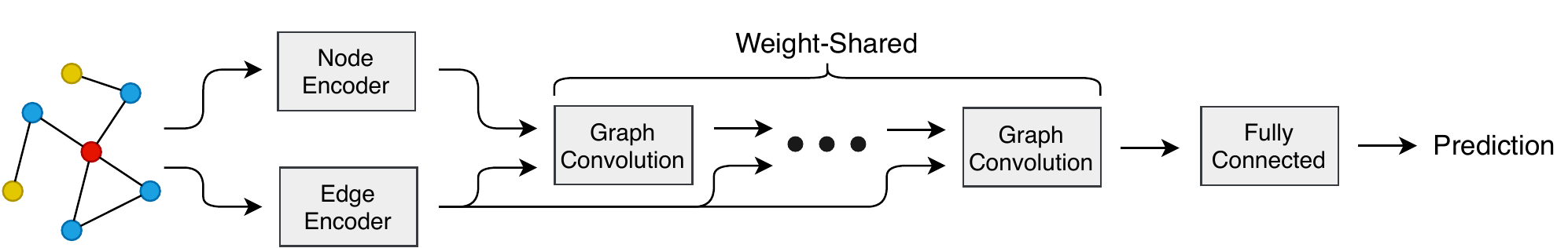}
    \caption{The full architecture of EWS-GNN with 3 graph convolutional layers. Graph convolutional layers share weights.}
    \label{fig:pipeline}
  \end{figure*}

  As discussed in Section~\ref{sec:proposed_layer}, the proposed convolutional layer preserves the dimension of node vector representation. This property allows us to use this part of a pipeline recurrently, running the cell several times. We underline that in all the experiments we take the same weight matrices \(W_E\) and \(W_X\) as well as parameters of the GRU cell \(g\) for all the layers. The recurrent usages helps to improve the representation power of the network while achieving the strong reduction in the number of trained weights. Finally, in order to predict the target, the last obtained embedding is passed to a linear layer, see the full pipeline on Figure~\ref{fig:pipeline}.

\section{Experiments}
\label{sec:experiments}

\subsection{Data Preprocessing}
  We build train and test datasets so that they do not intersect in target clients and time.
  Moreover, they have three-month gap in-between that allows to estimate model generalization more reliably.
  Train and test datasets contain \numprint{500000} and \numprint{50000} loan applications with $10\%$ and $20\%$ default rates respectively.
  Subsequently, we build the following datasets which depend on the different types of information available.
    
    \subsubsection{Each sample is based on a single client's information}
    \textit{ }
    \textbf{Raw transactional data}. In that case, we operate with raw transactional data of the client, as shown in Table~\ref{tab:transactions}. It can be directly processed by RNN-based methods.
     
    \textbf{Generated features based on transactional data}. In order to apply standard ML models like ensembles of trees we generate numerical features.
      Every transaction has its type (MCC code), amount, currency, date and time. So we can calculate sum/average/median/min/max transaction amount within any time period (7 groups), MCC group (40) and currency (3). 
      Also, we add different ratios of features obtained on the previous steps giving as a result more than 7'000 features.

    \subsubsection{Each sample is based on subgraph representing the neighborhood of the target client}
    \textit{ }

    \textbf{Raw subgraph data}. 
    We consider the neighborhoods of depth two and corresponding sequential data for both edges and nodes (as in Table~\ref{tab:transactions}).

    \textbf{Generated features based on transactional graph}.  
      We start by choosing the best 1000 features (according to the feature importance) which correspond to the target client
      and add features corresponding to the graph. First, we filter the target node neighbors by weighting the edges in different ways and by subsequent thresholding. Filter examples include the number and/or sum of transfers
      (35 filters), 
      and the number and/or sum of transfers within time period
      (175 filters). Thresholding removes part of the edges and we consider the remaining part as an unweighted graph. Then for the obtained filtered graph we compute different statistics (min/max/average/etc.) of PageRank, node degree, and betweenness centrality of edges. These operations give us more than 500 new features and we finally get more than 1'500 features.

\subsection{Graph Neural Networks Training Pipeline}
\label{sec:gnn_training}
  We propose a special training procedure for GNNs in order to get more stable training process, leading to better performance. First, we augment training dataset as follows:
  \begin{itemize}
    \item We randomly sample neighbors (20 to 25 nodes) and neighbors of neighbors (15 to 20 nodes). Thus, the subgraph consists of 500 nodes at maximum.

    \item We randomly delete transactions (0 to 25\%).
  \end{itemize}
  These augmentations improve the final quality of each GNN model, allowing using of a batch size equal to 16.
    
  We propose the following training procedure:
  \begin{enumerate}
    \item Node encoder initialization and edge encoder initialization (if exists) based on transactional neural network. We freeze these modules' weights for several epochs, training only graph modules.
  
    \item If edge encoder exists, we defreeze its weights, and continue training for several more epochs.
  
    \item We defreeze node encoder's weights and train the whole model.
  \end{enumerate}
    
    
  At each of the training stages, the StepLR scheduler was used with the initial value of learning rate equal to \(0.001\).
  Also, we use dropout for hidden layers, obtaining the best performance with dropout rate \(p=0.25\). 

  \begin{table*}[h]
  \centering
    \caption{Models comparison (ROC AUC).
    Pretraining helps to improve the quality compared to the random initialization and additional end-to-end training allows to achieve even better results. The EWS-GCN model benefits from the end-to-end training the most.} 
    \label{tab:model_train}
    \begin{tabular}{|l|ccc|}
      \hline
      Model & Random Initialization of Encoders & Frozen Pretrained Encoders (Stage 2) & End-to-End Training (Stage 3) \\
      \hline
      GAT & \(0.8517 \pm 0.0018\) & \(0.8842 \pm 0.0015\) & \(0.8875 \pm 0.0015\) \\
      EGNN & \(0.8655 \pm 0.0017\) & \(0.8849 \pm 0.0015\) & \(0.8885 \pm 0.0015\)\\
      EWS-GCN & \(0.8620 \pm 0.0017\) & \(0.8878 \pm 0.0015\) & \(\textbf{0.8926} \pm 0.0014\) \\
      \hline
    \end{tabular}
  \end{table*}

\subsection{Experimental Setup}
  We run all the experiments on the cluster of 8 PCs with the following configuration: 32 CPU cores, 256 GB RAM, Tesla P100 with 16 GB VRAM. We consider the following models.

\subsubsection{Models based on single client transactions}
  \textit{ }

  \textbf{Gradient Boosting Model.} As a base gradient boosting model, we take LightGBM~\cite{Ke2017} and train it on generated features based on transactional data. 
    
  \textbf{Transactional Neural Network}. As a transactional neural network model we take E.T.-RNN~\cite{Babaev2019}. The model is trained on raw transactional data. The implementation details are as follows:
    
  \begin{itemize} 
    \item All categorical features are preprocessed with embedding layers, while the only numerical feature (the sum of transactions) is normalized with batch normalization.
            
            
    \item As a base cell, GRU with 60 neurons in a hidden layer is chosen. The training starts with learning rate equal to 0.01 which is further adjusted by StepLR scheduler. 
            
  \end{itemize}

\subsubsection{Models based on subgraph data of the client}
  \textit{ }
    
  \textbf{Gradient Boosting Model on Transactional Graph.}
  For transactional graph, we as well take LightGBM as a base gradient boosting model and train it on generated features. The training details are as follows: 1000 iterations, maximum depth of 5 and learning rate equal to \(0.01\).
    
  \textbf{GAT.}
  We take implementation from \textit{Pytorch Geometric}~\cite{Fey2019}. This method does not take into account edge features, so we skip the second stage of the pipeline. The optimal performance is obtained on configuration of two convolutions. The first convolution has 8 heads, ELU activation, and is based on concatenation operation. The second one has 4 heads, ELU activation, and is based on the sum operation. Also, we concatenate all the intermediate representations obtained from convolutions, including the initial one, and pass the resulting vector into the fully connected layer. 

  \textbf{EGNN.}
  EGNN is trained based on edge features, so the training consists of all the pipeline steps from the previous section.
  We take convolutional layers with 8 aggregations, and ELU is used as the activation function for hidden layers. Similarly to GAT, we concatenate all the intermediate representations and use the resulting vector for the final prediction.

    
  \textbf{EWS-GCN.}
  EWS-GCN training consists of all three stages of the pipeline described in Section~\ref{sec:gnn_training}, since the model processes edge attributes. The hidden size dimension is equal to 80, while the initial vectors size is 60 (the output dimension of RNN-based encoder). One of the distinctive properties of the model is that graph convolutions are used recurrently, allowing increasing the depth of the network along with reducing the number of trainable weights. Thus, the model requires 2.1 times less VRAM compared to GAT model. 

\subsection{Results}
  We start the analysis by showing the benefits of the proposed training pipeline. Models improvement during the training and comparison with the training from random-initialization of parameters is demonstrated in Table~\ref{tab:model_train}. Standard errors were computed according to~\cite{cortes2005confidence}. The results show that pretraining helps to improve quality for all the models, while final end-to-end training improves the results even further.

\begin{figure}[h]
    \centering
    \includegraphics[width=0.8\linewidth]{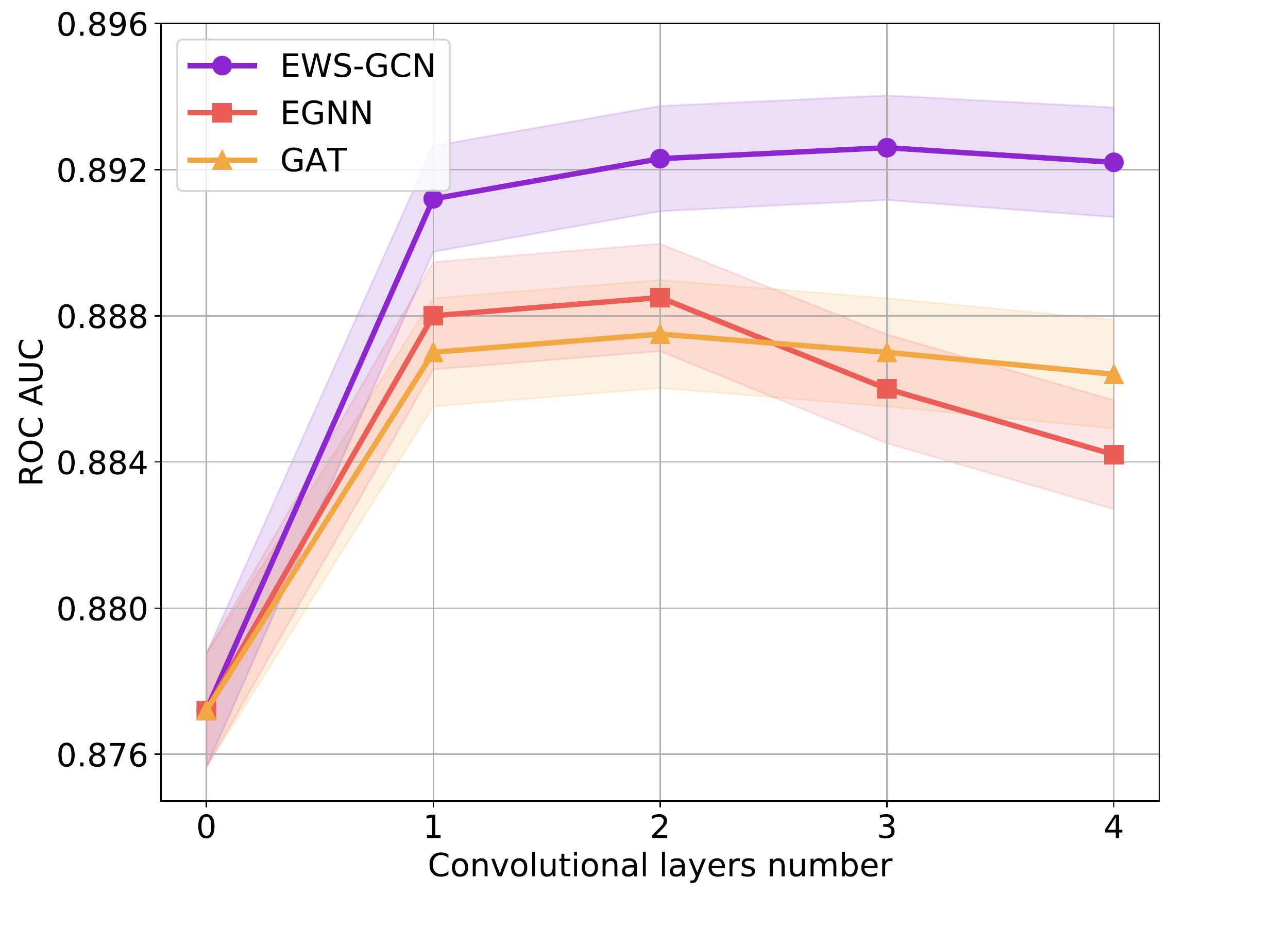}
    \caption{Comparison of GNN architectures scores as function of convolutional layers number. Point zero corresponds to E.T.-RNN (i.e., no usage of graph data).}
    \label{fig:conv_layer_graph}
  \end{figure}

  We additionally explore the dependence of the performance of GNNs on the number of convolutional layers. As you can see on the Figure~\ref{fig:conv_layer_graph}, the best score is achieved by EWS-GCN with three convolutions, in contrast to the competing methods which show the best results for the depth equal to two. 
    
  The final scores of the models discussed above can be found in Table~\ref{tab:model_data}. We see the strong superiority of the graph based approaches, and our proposed EWS-GCN model outperforms all the competitors. We should note that, while the considered difference between best performing models is just 0.004 AUC, it is still significant for the bank having millions of clients. 

  \begin{table}[h]
  \centering
    \caption{Models comparison. For GNNs we take the number of convolutional layers which gives best results, see also Figure~\ref{fig:conv_layer_graph}.}
    \label{tab:model_data}
    \begin{tabular}{|l|l|c|}
      \hline
      Model & Data & ROC AUC \\
      \hline
      LightGBM & Transactions & \(0.8638 \pm 0.0017\) \\
      E.T.-RNN & (1 client)  & \(0.8775 \pm 0.0016\) \\
      \hline
      LightGBM & Graph with & \(0.8830 \pm 0.0015\) \\
      GAT &  node attributes & \(0.8875 \pm 0.0015\) \\
      \hline
      EGNN & Graph with & \(0.8885 \pm 0.0015\) \\
      EWS-GCN & node and edge attributes & \(\textbf{0.8926} \pm 0.0014\) \\
      \hline
    \end{tabular}
  \end{table}

\section{Conclusion}
\label{sec:conclusions}
  Based on the framework of graph neural networks, we achieve high-quality prediction in the problem of the client credit scoring, significantly outperforming the classic ML approaches based on features available for the particular client. In addition, we demonstrate a complete sequence of models improving with enriching the data: client transactions, then a graph of transactions with information corresponding to nodes, and, finally, a graph of transactions with information both in nodes and edges. Additionally, we show the importance of improving the architecture of the neural network by considering GAT and EGNN, and proposing a new model EWS-GCN. In the pipeline we develop, we managed to achieve a significant increase in quality with attention mechanism based on information obtained not only from neighboring nodes attributes but also from edge attributes. Moreover, we show that the optimization of the number of weights used (graph convolutional layers share the weights) allows building deeper and more powerful architectures.


\bibliographystyle{IEEEtran}
\bibliography{bibliography}

\end{document}